\documentclass[runningheads]{llncs}
\usepackage{graphicx}
\usepackage{amsmath,amssymb} 
\usepackage{color}
\usepackage{multirow}
\usepackage[width=122mm,left=12mm,paperwidth=146mm,height=193mm,top=12mm,paperheight=217mm]{geometry}

\usepackage[bookmarks=false,colorlinks=true,linkcolor=black,citecolor=black,filecolor=black]{hyperref}
\usepackage{url}
\usepackage[british,UKenglish,USenglish,english,american]{babel}

\def\eg{\emph{e.g.}}

\def\ie{\emph{i.e.}}

\def\vs{\emph{vs. }}

\newcommand{\tabincellsize}[4]{\fontsize{#1}{#2em}\selectfont\begin{tabular}{@{}#3@{}}#4\end{tabular}}
\newcommand{\ve}[1]{\mathbf{#1}} 
\newcommand{\ma}[1]{\mathrm{#1}} 

\newcommand{\fp}[2]{\frac{\partial{#1}}{\partial{#2}}} 
\newcommand{\dvp}[2]{\frac{\partial}{\partial{#2}}#1} 

\begin{document}

\pagestyle{headings}
\mainmatter

\title{Identity Mappings in Deep Residual Networks}

\titlerunning{~}

\authorrunning{~}

\author{Kaiming He \and Xiangyu Zhang \and Shaoqing Ren \and Jian Sun}
\institute{Microsoft Research}

\maketitle

\begin{abstract}
Deep residual networks \cite{He2016} have emerged as a family of extremely deep architectures showing compelling accuracy and nice convergence behaviors. In this paper, we analyze the propagation formulations behind the residual building blocks, which suggest that the forward and backward signals can be directly propagated from one block to any other block, when using identity mappings as the skip connections and after-addition activation. A series of ablation experiments support the importance of these identity mappings. This motivates us to propose a new residual unit, which makes training easier and improves generalization. We report improved results using a 1001-layer ResNet on CIFAR-10 (4.62\% error) and CIFAR-100,  and a 200-layer ResNet on ImageNet. Code is available at: \url{https://github.com/KaimingHe/resnet-1k-layers}.
\end{abstract}


\hypersetup{urlcolor=black}

\section{Introduction}
\begin{figure}[t]
\centering
\includegraphics[width=1.0\linewidth]{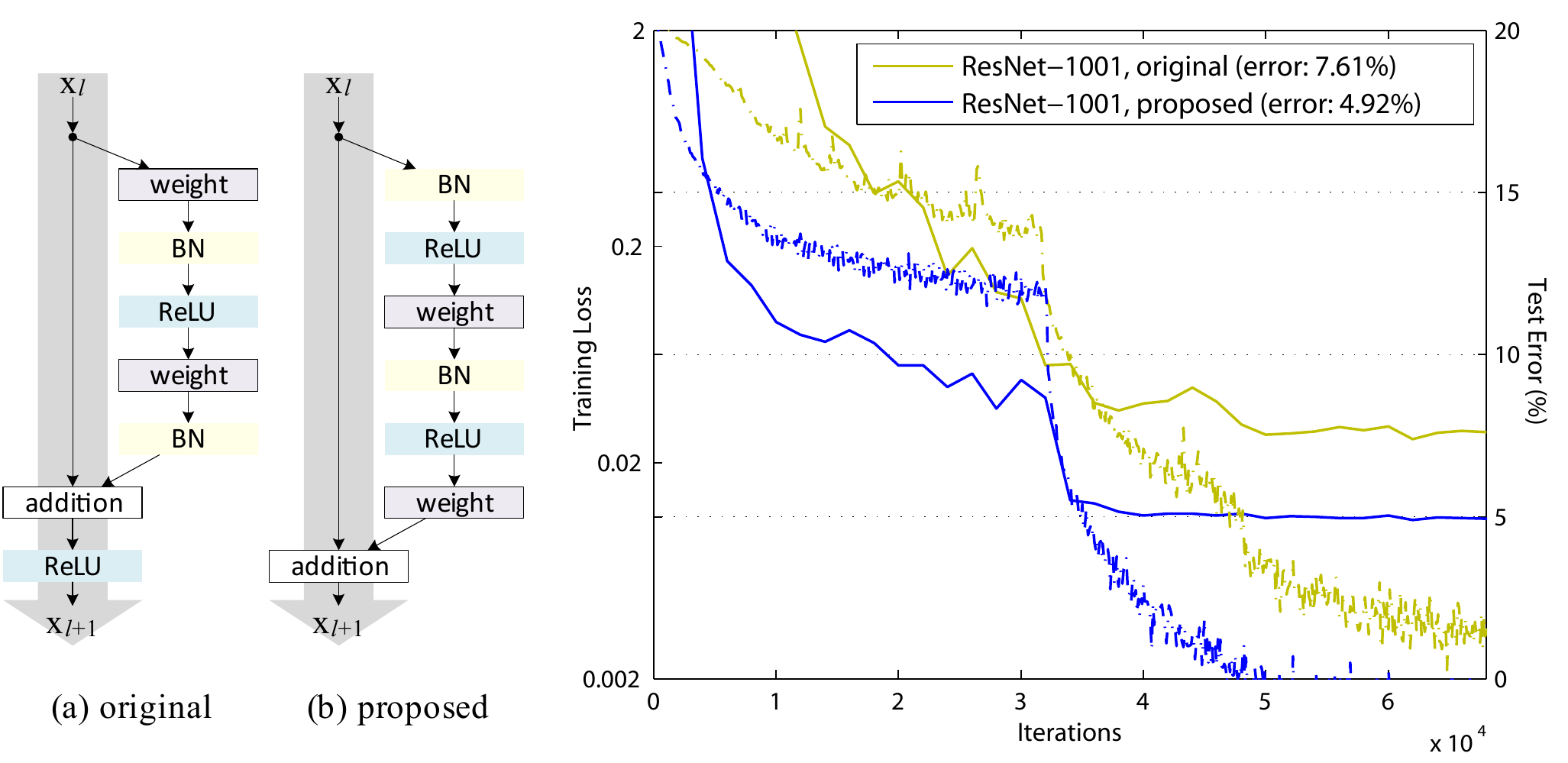}
\caption{\textbf{Left}: (a) original Residual Unit in \cite{He2016}; (b) proposed Residual Unit. The grey arrows indicate the easiest paths for the information to propagate, corresponding to the additive term ``$\ve{x}_l$'' in Eqn.(\ref{eq:additive}) (forward propagation) and the additive term ``1'' in Eqn.(\ref{eq:grad}) (backward propagation). \textbf{Right}: training curves on CIFAR-10 of \textbf{1001-layer} ResNets. Solid lines denote test error (y-axis on the right), and dashed lines denote training loss (y-axis on the left). The proposed unit makes ResNet-1001 easier to train.}
\label{fig:teaser}
\end{figure}

Deep residual networks (ResNets) \cite{He2016} consist of many stacked ``Residual Units". Each unit (Fig.~\ref{fig:teaser} (a)) can be expressed in a general form:
\begin{gather}
\ve{y}_{l} = h(\ve{x}_{l}) + \mathcal{F}(\ve{x}_{l}, \mathcal{W}_l), \nonumber\\
\ve{x}_{l+1} = f(\ve{y}_{l}) \nonumber,
\end{gather}
where $\ve{x}_{l}$ and $\ve{x}_{l+1}$ are input and output of the $l$-th unit, and $\mathcal{F}$ is a residual function. In \cite{He2016}, $h(\ve{x}_{l}) = \ve{x}_{l}$ is an identity mapping and $f$ is a ReLU \cite{Nair2010} function.

ResNets that are over 100-layer deep have shown state-of-the-art accuracy for several challenging recognition tasks on ImageNet \cite{Russakovsky2015} and MS COCO \cite{Lin2014} competitions. The central idea of ResNets is to learn the additive residual function $\mathcal{F}$ with respect to $h(\ve{x}_{l})$, with a key choice of using an identity mapping $h(\ve{x}_{l}) = \ve{x}_{l}$. This is realized by attaching an identity skip connection (``shortcut'').

In this paper, we analyze deep residual networks by focusing on creating a ``direct" path for propagating information --- not only within a residual unit, but through the entire network. Our derivations reveal that \emph{if both $h(\ve{x}_{l})$ and $f(\ve{y}_{l})$ are identity mappings}, the signal could be \emph{directly} propagated from one unit to any other units, in both forward and backward passes. Our experiments empirically show that training in general becomes easier when the architecture is closer to the above two conditions.

To understand the role of skip connections, we analyze and compare various types of $h(\ve{x}_{l})$. We find that the identity mapping $h(\ve{x}_{l}) = \ve{x}_{l}$ chosen in \cite{He2016} achieves the fastest error reduction and lowest training loss among all variants we investigated, whereas skip connections of scaling, gating \cite{Hochreiter1997,Srivastava2015,Srivastava2015a}, and 1$\times$1 convolutions all lead to higher training loss and error. These experiments suggest that keeping a ``clean'' information path (indicated by the grey arrows in Fig.~\ref{fig:teaser},~\ref{fig:shortcuts}, and~\ref{fig:activations}) is helpful for easing optimization.

To construct an identity mapping $f(\ve{y}_{l})=\ve{y}_{l}$, we view the activation functions (ReLU and BN~\cite{Ioffe2015}) as ``\emph{pre-activation}'' of the weight layers, in contrast to conventional wisdom of ``post-activation''. This point of view leads to a new residual unit design, shown in (Fig.~\ref{fig:teaser}(b)). Based on this unit, we present competitive results on CIFAR-10/100 with a 1001-layer ResNet, which is much easier to train and generalizes better than the original ResNet in \cite{He2016}. We further report improved results on ImageNet using a 200-layer ResNet, for which the counterpart of \cite{He2016} starts to overfit. These results suggest that there is much room to exploit the dimension of \emph{network depth}, a key to the success of modern deep learning.

\section{Analysis of Deep Residual Networks}

The ResNets developed in \cite{He2016} are \emph{modularized} architectures that stack building blocks of the same connecting shape. In this paper we call these blocks ``\emph{Residual Units}''. The original Residual Unit in \cite{He2016} performs the following computation:
\begin{gather}
\ve{y}_{l} = h(\ve{x}_{l}) + \mathcal{F}(\ve{x}_{l}, \mathcal{W}_l), \label{eq:resunit1}\\
\ve{x}_{l+1} = f(\ve{y}_{l}) \label{eq:resunit2}.
\end{gather}
Here $\ve{x}_{l}$ is the input feature to the $l$-th Residual Unit. $\mathcal{W}_l=\{\ma{W}_{l,k} | _{1\leq k \leq K}\}$ is a set of weights (and biases) associated with the $l$-th Residual Unit, and $K$ is the number of layers in a Residual Unit ($K$ is 2 or 3 in \cite{He2016}). $\mathcal{F}$ denotes the residual function, \eg, a stack of two 3$\times$3 convolutional layers in \cite{He2016}. The function $f$ is the operation after element-wise addition, and in \cite{He2016} $f$ is ReLU. The function $h$ is set as an identity mapping: $h(\ve{x}_{l}) = \ve{x}_{l}$.\footnote{It is noteworthy that there are Residual Units for increasing dimensions and reducing feature map sizes \cite{He2016} in which $h$ is not identity.
In this case the following derivations do not hold strictly. But as there are only a very few such units (two on CIFAR and three on ImageNet, depending on image sizes \cite{He2016}), we expect that they do not have the exponential impact as we present in Sec.~\ref{sec:identityskip}. One may also think of our derivations as applied to all Residual Units within the same feature map size.}

If $f$ is also an identity mapping: $\ve{x}_{l+1} \equiv \ve{y}_{l}$, we can put Eqn.(\ref{eq:resunit2}) into Eqn.(\ref{eq:resunit1}) and obtain:
\begin{equation}
\ve{x}_{l+1} = \ve{x}_{l} + \mathcal{F}(\ve{x}_{l}, \mathcal{W}_{l}). \label{eq:additive0}
\end{equation}
Recursively ({\fontsize{8pt}{1em}$\ve{x}_{l+2} = \ve{x}_{l+1} + \mathcal{F}(\ve{x}_{l+1},\mathcal{W}_{l+1})=\ve{x}_{l} + \mathcal{F}(\ve{x}_{l}, \mathcal{W}_{l})+\mathcal{F}(\ve{x}_{l+1}, \mathcal{W}_{l+1})$}, etc.) we will have:
\begin{equation}
\ve{x}_{L} = \ve{x}_{l} + \sum_{i=l}^{L-1}\mathcal{F}(\ve{x}_{i}, \mathcal{W}_{i}), \label{eq:additive}
\end{equation}
for \emph{any deeper unit} $L$ and \emph{any shallower unit} $l$. Eqn.(\ref{eq:additive}) exhibits some nice properties. \textbf{(i)} The feature $\ve{x}_L$ of any deeper unit $L$ can be represented as the feature $\ve{x}_l$ of any shallower unit $l$ plus a residual function in a form of $\sum_{i=l}^{L-1}\mathcal{F}$, indicating that the model is in a \emph{residual} fashion between any units $L$ and $l$. \textbf{(ii)} The feature $\ve{x}_{L} = \ve{x}_{0} + \sum_{i=0}^{L-1}\mathcal{F}(\ve{x}_{i}, \mathcal{W}_{i})$, of any deep unit $L$,  is the \emph{summation} of the outputs of all preceding residual functions (plus $\ve{x}_{0}$). This is in contrast to a ``plain network'' where a feature $\ve{x}_{L}$ is a series of matrix-vector \emph{products}, say, $\prod_{i=0}^{L-1}W_{i}\ve{x}_0$ (ignoring BN and ReLU).

Eqn.(\ref{eq:additive}) also leads to nice backward propagation properties.
Denoting the loss function as $\mathcal{E}$, from the chain rule of backpropagation \cite{LeCun1989} we have:
\begin{equation}
\fp{\mathcal{E}}{\ve{x}_{l}}=\fp{\mathcal{E}}{\ve{x}_{L}}\fp{\ve{x}_{L}}{\ve{x}_{l}}=\fp{\mathcal{E}}{\ve{x}_{L}}\left(1+\dvp{\sum_{i=l}^{L-1}\mathcal{F}(\ve{x}_{i}, \mathcal{W}_{i})}{\ve{x}_{l}}\right).\label{eq:grad}
\end{equation}
Eqn.(\ref{eq:grad}) indicates that the gradient $\fp{\mathcal{E}}{\ve{x}_{l}}$ can be decomposed into two additive terms: a term of $\fp{\mathcal{E}}{\ve{x}_{L}}$ that propagates information directly without concerning any weight layers, and another term of {\fontsize{8pt}{1em}\selectfont $\fp{\mathcal{E}}{\ve{x}_{L}}\left(\dvp{\sum_{i=l}^{L-1}\mathcal{F}}{\ve{x}_{l}}\right)$} that propagates through the weight layers. The additive term of $\fp{\mathcal{E}}{\ve{x}_{L}}$ ensures that information is directly propagated back to \emph{any shallower unit} $l$.
Eqn.(\ref{eq:grad}) also suggests that it is unlikely for the gradient $\fp{\mathcal{E}}{\ve{x}_{l}}$ to be canceled out for a mini-batch, because in general the term {\fontsize{8pt}{1em}\selectfont $\dvp{\sum_{i=l}^{L-1}\mathcal{F}}{\ve{x}_{l}}$} cannot be always -1 for all samples in a mini-batch. This implies that the gradient of a layer does not vanish even when the weights are arbitrarily small.

\subsection*{Discussions}

Eqn.(\ref{eq:additive}) and Eqn.(\ref{eq:grad}) suggest that the signal can be directly propagated from any unit to another, both forward and backward.
The foundation of Eqn.(\ref{eq:additive}) is two identity mappings: (i) the identity skip connection $h(\ve{x}_{l}) = \ve{x}_{l}$, and (ii) the condition that $f$ is an identity mapping.

\begin{figure}[t]
\centering
\includegraphics[width=.7\linewidth]{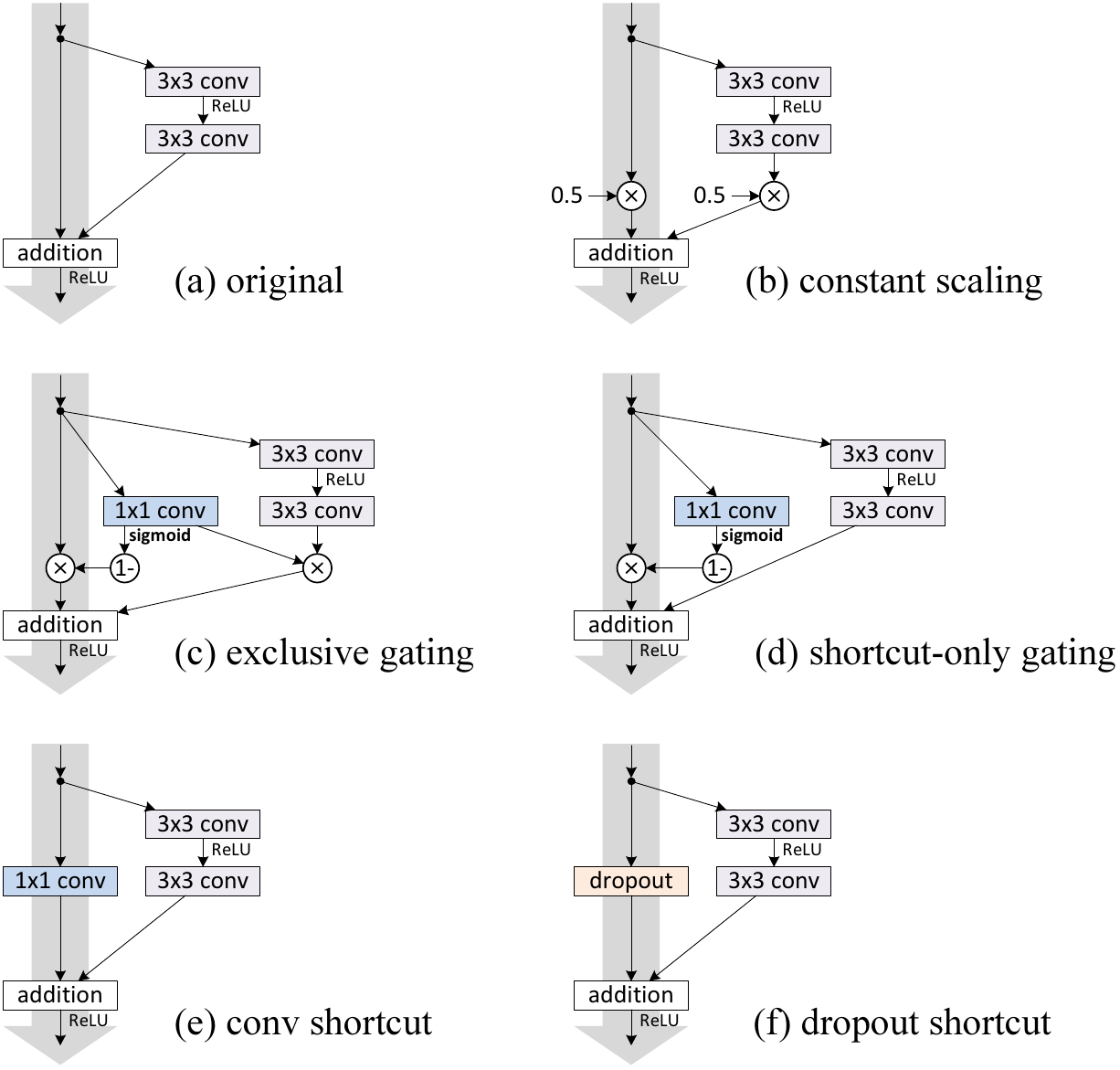}
\caption{Various types of shortcut connections used in Table~\ref{tab:shortcuts}. The grey arrows indicate the easiest paths for the information to propagate. The shortcut connections in (b-f) are impeded by different components. For simplifying illustrations we do not display the BN layers, which are adopted right after the weight layers for all units here.}
\label{fig:shortcuts}
\end{figure}

These directly propagated information flows are represented by the grey arrows in Fig.~\ref{fig:teaser},~\ref{fig:shortcuts}, and~\ref{sec:activations}. And the above two conditions are true when these grey arrows cover no operations (expect addition) and thus are ``clean''.
In the following two sections we separately investigate the impacts of the two conditions.


\section{On the Importance of Identity Skip Connections}
\label{sec:identityskip}

Let's consider a simple modification, $h(\ve{x}_{l}) = \lambda_l\ve{x}_{l}$, to break the identity shortcut:
\begin{eqnarray}
\ve{x}_{l+1} = \lambda_l\ve{x}_{l} + \mathcal{F}(\ve{x}_{l}, \mathcal{W}_{l}),
\end{eqnarray}
where $\lambda_l$ is a modulating scalar (for simplicity we still assume $f$ is identity). Recursively applying this formulation we obtain an equation similar to Eqn.~(\ref{eq:additive}): $\ve{x}_{L} =  (\prod_{i=l}^{L-1}\lambda_{i})\ve{x}_{l} + \sum_{i=l}^{L-1} (\prod_{j=i+1}^{L-1}\lambda_{\tiny j}) \mathcal{F}(\ve{x}_{i}, \mathcal{W}_{i})$, or simply:
\begin{eqnarray}
\ve{x}_{L} = (\prod_{i=l}^{L-1}\lambda_{i})\ve{x}_{l} + \sum_{i=l}^{L-1}\mathcal{\hat{F}}(\ve{x}_{i}, \mathcal{W}_{i}),
\label{eq:additive1}
\end{eqnarray}
where the notation $\mathcal{\hat{F}}$ absorbs the scalars into the residual functions.
Similar to Eqn.(\ref{eq:grad}), we have backpropagation of the following form:
\begin{equation}
\fp{\mathcal{E}}{\ve{x}_{l}}=\fp{\mathcal{E}}{\ve{x}_{L}}\left((\prod_{i=l}^{L-1}\lambda_{i})+\dvp{\sum_{i=l}^{L-1}\mathcal{\hat{F}}(\ve{x}_{i}, \mathcal{W}_{i})}{\ve{x}_{l}}\right).\label{eq:grad1}
\end{equation}
Unlike Eqn.(\ref{eq:grad}), in Eqn.(\ref{eq:grad1}) the first additive term is modulated by a factor $\prod_{i=l}^{L-1}\lambda_{i}$. For an extremely deep network ($L$ is large), if $\lambda_{i}>1$ for all $i$, this factor can be exponentially large; if $\lambda_{i}<1$ for all $i$, this factor can be exponentially small and vanish, which blocks the backpropagated signal from the shortcut and forces it to flow through the weight layers. This results in optimization difficulties as we show by experiments.

In the above analysis, the original identity skip connection in Eqn.(\ref{eq:additive0}) is replaced with a simple scaling $h(\ve{x}_{l}) = \lambda_l\ve{x}_{l}$. If the skip connection $h(\ve{x}_{l})$ represents more complicated transforms (such as gating and 1$\times$1 convolutions), in Eqn.(\ref{eq:grad1}) the first term becomes $\prod_{i=l}^{L-1}h'_{i}$ where $h'$ is the derivative of $h$. This product may also impede information propagation and hamper the training procedure as witnessed in the following experiments.

\renewcommand\arraystretch{1.3}
\newcommand{\bsize}[1]{\fontsize{#1pt}{1em}\selectfont}
\setlength{\tabcolsep}{6pt}
\begin{table}[t]
\caption{Classification error on the CIFAR-10 test set using ResNet-110 \cite{He2016}, with different types of shortcut connections applied to all Residual Units. We report ``fail'' when the test error is higher than 20\%.}\label{tab:shortcuts}
\centering
\fontsize{8pt}{1em}\selectfont
\begin{tabular}{c|l|c|c|c|l}
\hline
case & Fig. & on shortcut & on $\mathcal{F}$ & error (\%) & remark \\
\hline
\hline
original \cite{He2016} & Fig.~\ref{fig:shortcuts}(a) & 1 & 1  & \textbf{6.61} \\
\hline
\multirow{3}{*}{\tabincellsize{8}{.8}{c}{constant \\ scaling}}
& \multirow{3}{*}{Fig.~\ref{fig:shortcuts}(b)}
& 0 & 1 & {\bsize{8}fail} & \bsize{6} This is a plain net\\
&  & 0.5 & 1  & {\bsize{8}fail} \\
&  & 0.5 & 0.5 & 12.35 & {\bsize{6} frozen gating} \\
\hline
\multirow{3}{*}{\tabincellsize{8}{.8}{c}{exclusive \\ gating}}
& \multirow{3}{*}{Fig.~\ref{fig:shortcuts}(c)}
& $1-g(\ve{x})$ & $g(\ve{x})$  & {\bsize{8}fail} & \bsize{6} init $b_{g}$=0 to $-5$ \\
&  & $1-g(\ve{x})$ & $g(\ve{x})$  &  8.70                    & \bsize{6} init $b_{g}$=-6 \\
&  & $1-g(\ve{x})$ & $g(\ve{x})$  &  9.81                    & \bsize{6} init $b_{g}$=-7 \\
\hline
\multirow{2}{*}{\tabincellsize{8}{.8}{c}{shortcut-only \\ gating}}
& \multirow{2}{*}{Fig.~\ref{fig:shortcuts}(d)}
& $1-g(\ve{x})$ & 1  & 12.86 & \bsize{6} init $b_{g}$=0 \\
&  & $1-g(\ve{x})$ & 1  &  6.91 & \bsize{6} init $b_{g}$=-6 \\
\hline
1$\times$1 conv shortcut & \multirow{1}{*}{Fig.~\ref{fig:shortcuts}(e)}
& 1$\times$1 conv & 1  & 12.22 & \\
\hline
dropout shortcut & \multirow{1}{*}{Fig.~\ref{fig:shortcuts}(f)}
& dropout 0.5 & 1  & {\bsize{8}fail} & \\
\hline
\end{tabular}
\end{table}

\subsection{Experiments on Skip Connections}

We experiment with the 110-layer ResNet as presented in \cite{He2016} on CIFAR-10 \cite{Krizhevsky2009}. This extremely deep ResNet-110 has 54 two-layer Residual Units (consisting of 3$\times$3 convolutional layers) and is challenging for optimization.
Our implementation details (see appendix) are the same as \cite{He2016}. Throughout this paper we report the median accuracy of \textbf{5 runs} for each architecture on CIFAR, reducing the impacts of random variations.

Though our above analysis is driven by identity $f$, the experiments in this section are all based on $f=$ ReLU as in \cite{He2016}; we address identity $f$ in the next section.
Our baseline ResNet-110 has 6.61\% error on the test set.
The comparisons of other variants (Fig.~\ref{fig:shortcuts} and Table~\ref{tab:shortcuts}) are summarized as follows:

\begin{figure}[t]
\centering
\includegraphics[width=0.99\linewidth]{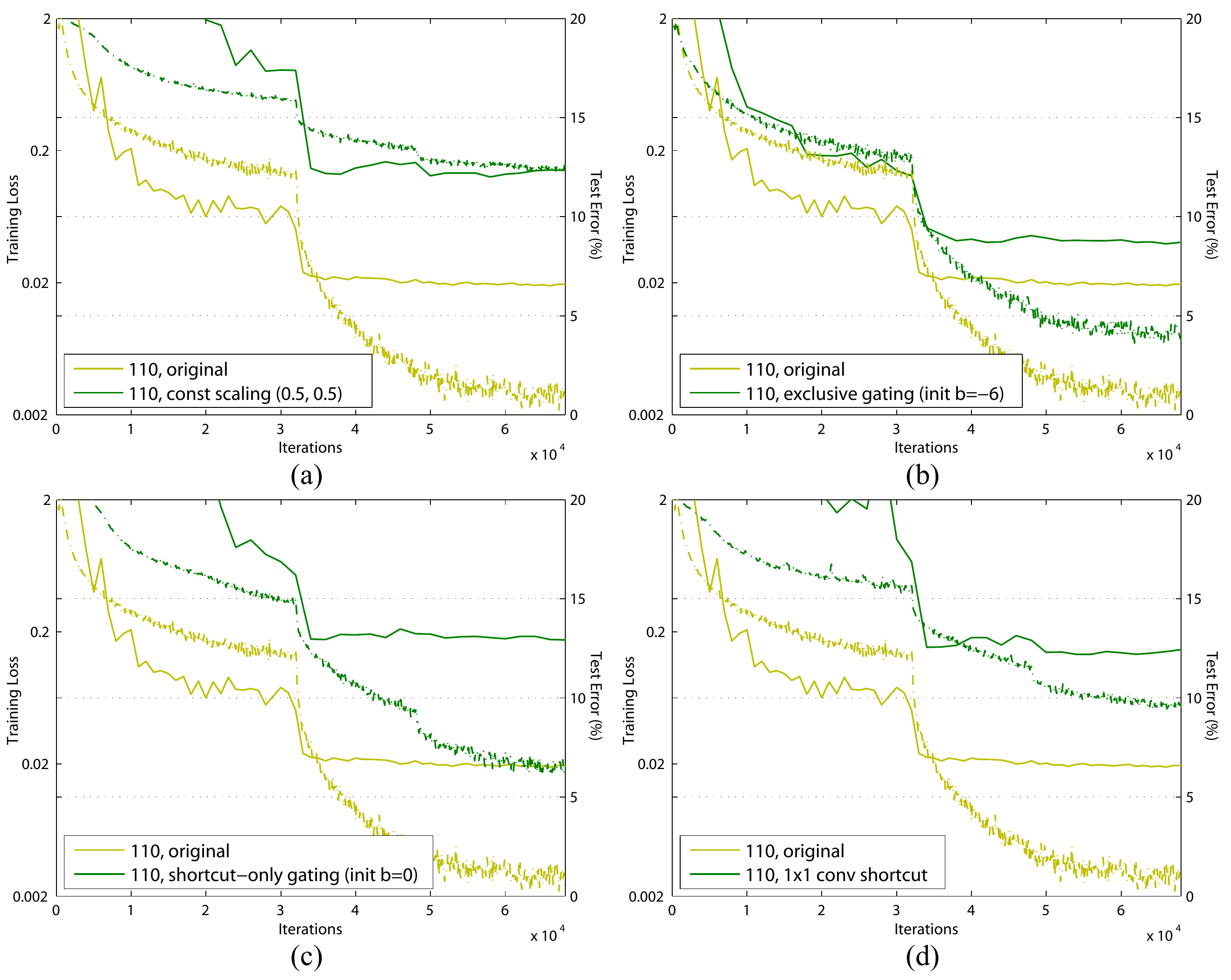}
\caption{Training curves on CIFAR-10 of various shortcuts. Solid lines denote test error (y-axis on the right), and dashed lines denote training loss (y-axis on the left).
}
\label{fig:curves_shortcuts}
\end{figure}

\textbf{Constant scaling}. We set $\lambda=0.5$ for all shortcuts (Fig.~\ref{fig:shortcuts}(b)). We further study two cases of scaling $\mathcal{F}$: (i) $\mathcal{F}$ is not scaled; or (ii) $\mathcal{F}$ is scaled by a constant scalar of $1-\lambda=0.5$, which is similar to the highway gating \cite{Srivastava2015,Srivastava2015a} but with frozen gates. The former case does not converge well; the latter is able to converge, but the test error (Table~\ref{tab:shortcuts}, 12.35\%) is substantially higher than the original ResNet-110. Fig~\ref{fig:curves_shortcuts}(a) shows that the training error is higher than that of the original ResNet-110, suggesting that the optimization has difficulties when the shortcut signal is scaled down.

\textbf{Exclusive gating}. Following the Highway Networks \cite{Srivastava2015,Srivastava2015a} that adopt a gating mechanism \cite{Hochreiter1997},
we consider a gating function $g(\ve{x})=\sigma(\ma{W}_g\ve{x}+b_g)$ where a transform is represented by weights $\ma{W}_g$ and biases $b_g$ followed by the sigmoid function $\sigma(x)=\frac{1}{1+e^{-x}}$. In a convolutional network $g(\ve{x})$ is realized by a 1$\times$1 convolutional layer. The gating function modulates the signal by element-wise multiplication.

We investigate the ``exclusive'' gates as used in \cite{Srivastava2015,Srivastava2015a} --- the $\mathcal{F}$ path is scaled by $g(\ve{x})$ and the shortcut path is scaled by $1-g(\ve{x})$. See Fig~\ref{fig:shortcuts}(c). We find that the initialization of the biases $b_g$ is critical for training gated models, and following the guidelines\footnote{See also: \url{people.idsia.ch/~rupesh/very_deep_learning/} by \cite{Srivastava2015,Srivastava2015a}.} in \cite{Srivastava2015,Srivastava2015a}, we conduct hyper-parameter search on the initial value of $b_g$ in the range of 0 to -10 with a decrement step of -1 on the training set by cross-validation. The best value ($-6$ here) is then used for training on the training set, leading to a test result of 8.70\% (Table~\ref{tab:shortcuts}), which still lags far behind the ResNet-110 baseline. Fig~\ref{fig:curves_shortcuts}(b) shows the training curves.
Table~\ref{tab:shortcuts} also reports the results of using other initialized values, noting that the exclusive gating network does not converge to a good solution when $b_g$ is not appropriately initialized.

The impact of the exclusive gating mechanism is two-fold. When $1-g(\ve{x})$ approaches 1, the gated shortcut connections are closer to identity which helps information propagation; but in this case $g(\ve{x})$ approaches 0 and suppresses the function $\mathcal{F}$. To isolate the effects of the gating functions on the shortcut path alone, we investigate a non-exclusive gating mechanism in the next.

\textbf{Shortcut-only gating}. In this case the function $\mathcal{F}$ is not scaled; only the shortcut path is gated by $1-g(\ve{x})$. See Fig~\ref{fig:shortcuts}(d). The initialized value of $b_g$ is still essential in this case. When the initialized $b_g$ is 0 (so initially the expectation of $1-g(\ve{x})$ is 0.5), the network converges to a poor result of 12.86\% (Table~\ref{tab:shortcuts}). This is also caused by higher training error (Fig~\ref{fig:curves_shortcuts}(c)).

When the initialized $b_g$ is very negatively biased (\eg, $-6$), the value of $1-g(\ve{x})$ is closer to 1 and the shortcut connection is nearly an identity mapping. Therefore, the result (6.91\%, Table~\ref{tab:shortcuts}) is much closer to the ResNet-110 baseline.

\textbf{1$\times$1 convolutional shortcut}. Next we experiment with 1$\times$1 convolutional shortcut connections that replace the identity. This option has been investigated in \cite{He2016} (known as option C) on a 34-layer ResNet (16 Residual Units) and shows good results, suggesting that 1$\times$1 shortcut connections could be useful. But we find that this is not the case when there are many Residual Units. The 110-layer ResNet has a poorer result (12.22\%, Table~\ref{tab:shortcuts}) when using 1$\times$1 convolutional shortcuts. Again, the training error becomes higher (Fig~\ref{fig:curves_shortcuts}(d)). When stacking so many Residual Units (54 for ResNet-110), even the shortest path may still impede signal propagation. We witnessed similar phenomena on ImageNet with ResNet-101 when using 1$\times$1 convolutional shortcuts.

\textbf{Dropout shortcut}. Last we experiment with dropout \cite{Hinton2012} (at a ratio of 0.5) which we adopt on the output of the identity shortcut (Fig.~\ref{fig:shortcuts}(f)). The network fails to converge to a good solution. Dropout statistically imposes a scale of $\lambda$ with an expectation of 0.5 on the shortcut, and similar to constant scaling by 0.5, it impedes signal propagation.

\subsection{Discussions}

As indicated by the grey arrows in Fig.~\ref{fig:shortcuts}, the shortcut connections are the most direct paths for the information to propagate. \emph{Multiplicative} manipulations (scaling, gating, 1$\times$1 convolutions, and dropout) on the shortcuts can hamper information propagation and lead to optimization problems.

It is noteworthy that the gating and 1$\times$1 convolutional shortcuts introduce more parameters, and should have stronger \emph{representational} abilities than identity shortcuts. In fact, the shortcut-only gating and 1$\times$1 convolution cover the solution space of identity shortcuts (\ie, they could be optimized as identity shortcuts). However, their training error is higher than that of identity shortcuts, indicating that the degradation of these models is caused by optimization issues, instead of representational abilities.

\section{On the Usage of Activation Functions}
\label{sec:activations}

\renewcommand\arraystretch{1.3}
\setlength{\tabcolsep}{6pt}
\begin{table}[t]
\caption{Classification error (\%) on the CIFAR-10 test set using different activation functions.}\label{tab:activations}
\centering
\fontsize{8pt}{1em}\selectfont
\begin{tabular}{l|l|c|c}
\hline
case & Fig. & ResNet-110 & ResNet-164 \\
\hline
\hline
original Residual Unit \cite{He2016} & Fig.~\ref{fig:activations}(a) & 6.61 & 5.93 \\
\hline
BN after addition & Fig.~\ref{fig:activations}(b) & 8.17 & 6.50 \\
\hline
ReLU before addition & Fig.~\ref{fig:activations}(c) & 7.84 & 6.14 \\
ReLU-only pre-activation & Fig.~\ref{fig:activations}(d) & 6.71 & 5.91 \\
\textbf{full pre-activation} & Fig.~\ref{fig:activations}(e) & \textbf{6.37} & \textbf{5.46} \\
\hline
\end{tabular}
\end{table}

\begin{figure}[t]
\centering
\includegraphics[width=.99\linewidth]{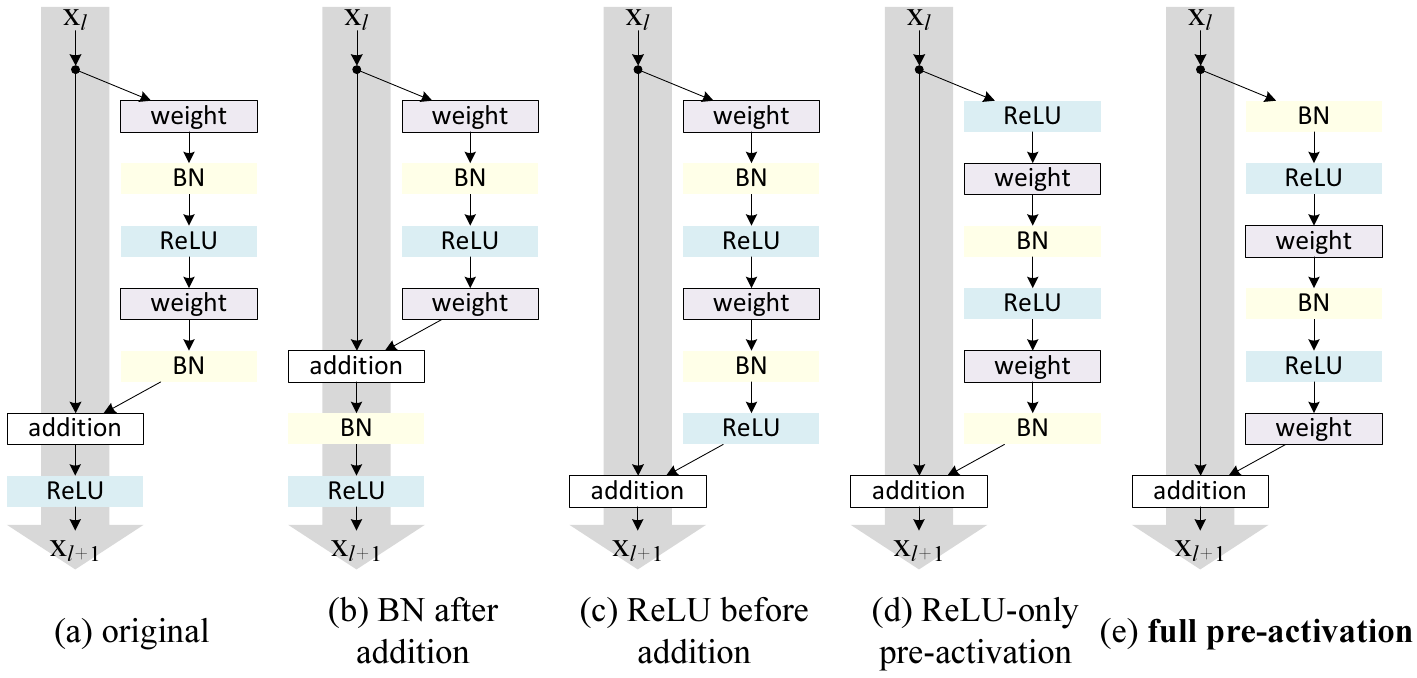}
\caption{Various usages of activation in Table~\ref{tab:activations}. All these units consist of the same components --- only the orders are different.}
\label{fig:activations}
\end{figure}

Experiments in the above section support the analysis in Eqn.(\ref{eq:grad}) and Eqn.(\ref{eq:grad1}), both being derived under the assumption that the after-addition activation $f$ is the identity mapping. But in the above experiments $f$ is ReLU as designed in \cite{He2016}, so Eqn.(\ref{eq:grad}) and (\ref{eq:grad1}) are approximate in the above experiments. Next we investigate the impact of $f$.

We want to make $f$ an identity mapping, which is done by re-arranging the activation functions (ReLU and/or BN). The original Residual Unit in \cite{He2016} has a shape in Fig.~\ref{fig:activations}(a) --- BN is used after each weight layer, and ReLU is adopted after BN except that the last ReLU in a Residual Unit is after element-wise addition ($f=$  ReLU). Fig.~\ref{fig:activations}(b-e) show the alternatives we investigated, explained as following.

\subsection{Experiments on Activation}

In this section we experiment with ResNet-110 and a 164-layer \emph{Bottleneck} \cite{He2016} architecture (denoted as ResNet-164). A bottleneck Residual Unit consist of a 1$\times$1 layer for reducing dimension, a 3$\times$3 layer, and a 1$\times$1 layer for restoring dimension. As designed in \cite{He2016}, its computational complexity is similar to the two-3$\times$3 Residual Unit. More details are in the appendix.
The baseline ResNet-164 has a competitive result of 5.93\% on CIFAR-10 (Table~\ref{tab:activations}).

\begin{figure}[t]
\centering
\includegraphics[width=0.99\linewidth]{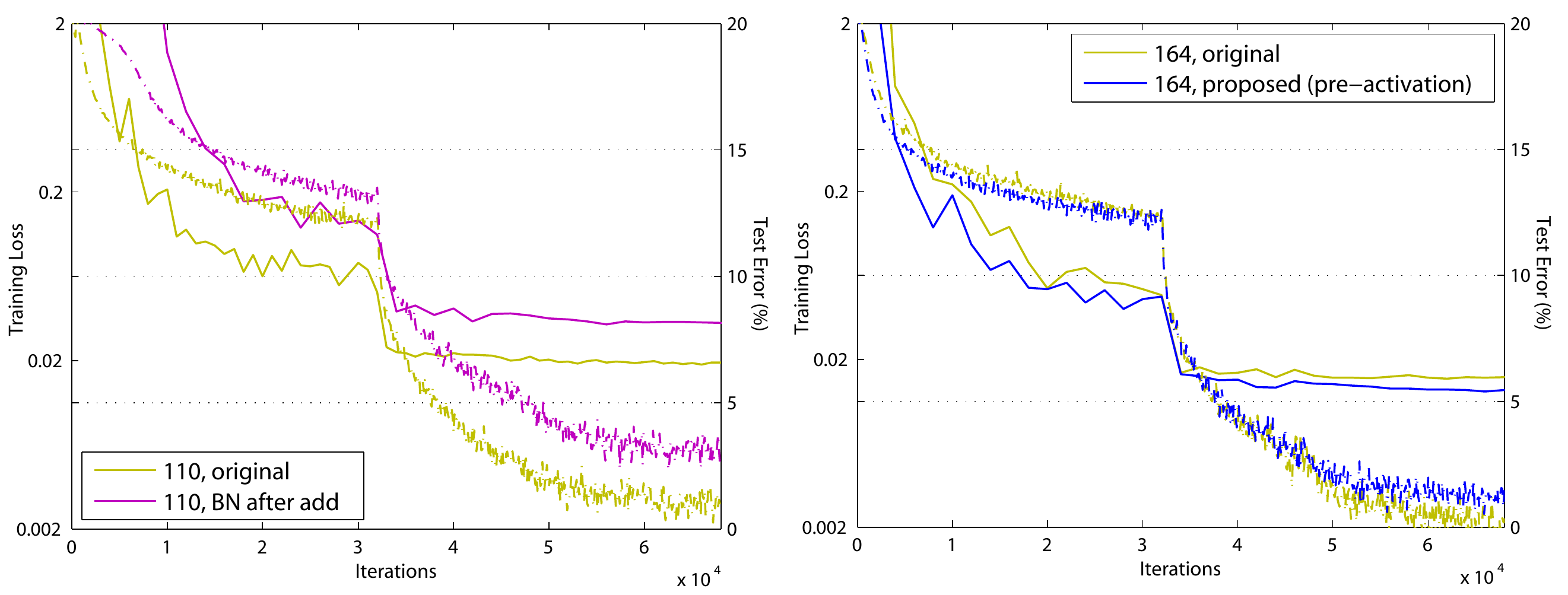}
\caption{Training curves on CIFAR-10. \textbf{Left}: BN after addition (Fig.~\ref{fig:activations}(b)) using ResNet-110. \textbf{Right}: pre-activation unit (Fig.~\ref{fig:activations}(e)) on ResNet-164. Solid lines denote test error, and dashed lines denote training loss.
}
\label{fig:curves_act}
\end{figure}

\textbf{BN after addition}. Before turning $f$ into an identity mapping, we go the opposite way by adopting BN after addition (Fig.~\ref{fig:activations}(b)). In this case $f$ involves BN and ReLU.
The results become considerably worse than the baseline (Table~\ref{tab:activations}). Unlike the original design, now the BN layer alters the signal that passes through the shortcut and impedes information propagation, as reflected by the difficulties on reducing training loss at the beginning of training (Fib.~\ref{fig:curves_act} left).

\textbf{ReLU before addition}. A na\"{i}ve choice of making $f$ into an identity mapping is to move the ReLU before addition (Fig.~\ref{sec:activations}(c)). However, this leads to a \emph{non-negative} output from the transform $\mathcal{F}$, while intuitively a ``residual'' function should take values in $(-\infty, +\infty)$. As a result, the forward propagated signal is monotonically increasing. This may impact the representational ability, and the result is worse (7.84\%, Table~\ref{tab:activations}) than the baseline.
We expect to have a residual function taking values in $(-\infty, +\infty)$. This condition is satisfied by other Residual Units including the following ones.

\begin{figure}[t]
\centering
\includegraphics[width=0.99\linewidth]{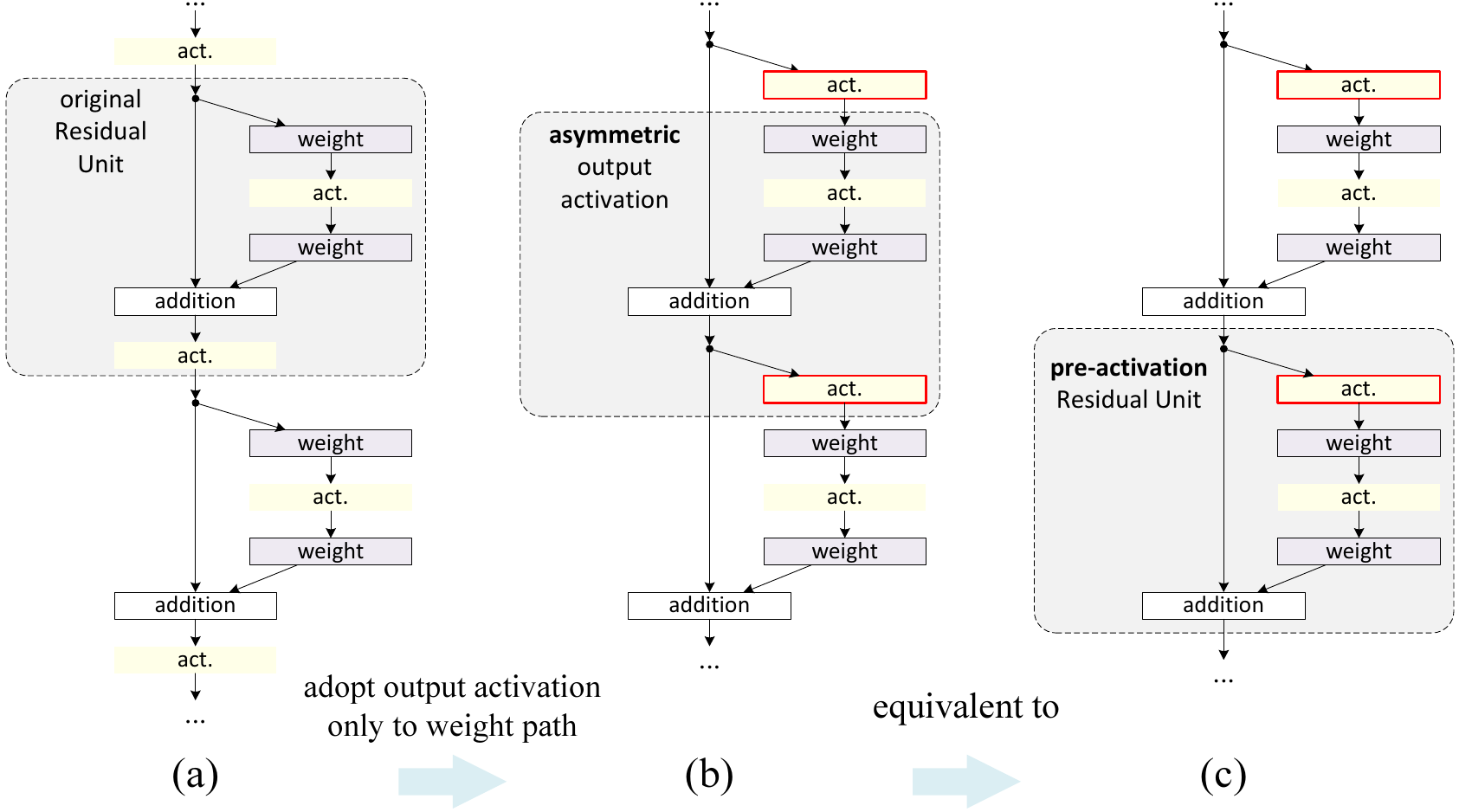}
\caption{Using asymmetric after-addition activation is equivalent to constructing a \emph{pre-activation} Residual Unit.}
\label{fig:asymmetric}
\end{figure}

\textbf{Post-activation or pre-activation?}
In the original design (Eqn.(\ref{eq:resunit1}) and Eqn.(\ref{eq:resunit2})), the activation $\ve{x}_{l+1}=f(\ve{y}_{l})$ affects \emph{both paths} in the \emph{next} Residual Unit: $\ve{y}_{l+1} = f(\ve{y}_{l}) + \mathcal{F}(f(\ve{y}_{l}), \mathcal{W}_{l+1})$.
Next we develop an \emph{asymmetric} form where an activation $\hat{f}$ only affects the $\mathcal{F}$ path: $\ve{y}_{l+1} = \ve{y}_{l} + \mathcal{F}(\hat{f}(\ve{y}_{l}), \mathcal{W}_{l+1})$, for any $l$ (Fig.~\ref{fig:asymmetric} (a) to (b)). By renaming the notations, we have the following form:
\begin{equation}
\ve{x}_{l+1} = \ve{x}_{l} + \mathcal{F}(\hat{f}(\ve{x}_{l}), \mathcal{W}_{l}), \label{eq:additive3}.
\end{equation}
It is easy to see that Eqn.(\ref{eq:additive3}) is similar to Eqn.(\ref{eq:additive}), and can enable a backward formulation similar to Eqn.(\ref{eq:grad}). For this new Residual Unit as in Eqn.(\ref{eq:additive3}), the new after-addition activation becomes an identity mapping.
This design means that if a new after-addition activation $\hat{f}$ is asymmetrically adopted, it is equivalent to recasting $\hat{f}$ as the \emph{pre-activation} of the next Residual Unit. This is illustrated in Fig.~\ref{fig:asymmetric}.

The distinction between post-activation/pre-activation is caused by the presence of the element-wise \emph{addition}. For a plain network that has $N$ layers, there are $N-1$ activations (BN/ReLU), and it does not matter whether we think of them as post- or pre-activations. But for branched layers merged by addition, the position of activation matters.

We experiment with two such designs: (i) ReLU-only pre-activation (Fig.~\ref{fig:activations}(d)), and (ii) full pre-activation (Fig.~\ref{fig:activations}(e)) where BN and ReLU are both adopted before weight layers. Table~\ref{tab:activations} shows that the ReLU-only pre-activation performs very similar to the baseline on ResNet-110/164. This ReLU layer is not used in conjunction with a BN layer, and may not enjoy the benefits of BN \cite{Ioffe2015}.

\renewcommand\arraystretch{1.3}
\setlength{\tabcolsep}{4pt}
\begin{table}[t]
\caption{Classification error (\%) on the CIFAR-10/100 test set using the original Residual Units and our pre-activation Residual Units.}\label{tab:preact}
\centering
\fontsize{8pt}{1em}\selectfont
\begin{tabular}{c|l|c|c}
\hline
dataset & network & baseline unit & pre-activation unit \\
\hline
\multirow{4}{*}{CIFAR-10}
& ResNet-110 (1layer skip) & 9.90 & \underline{8.91}\\
& ResNet-110 & 6.61 & \underline{6.37}\\
& ResNet-164 & 5.93 & \underline{5.46}\\
& ResNet-1001 & 7.61 & \underline{4.92}\\
\hline
\multirow{2}{*}{CIFAR-100}
& ResNet-164 & 25.16 & \underline{24.33}\\
& ResNet-1001 & 27.82  & \underline{22.71}\\
\hline
\end{tabular}
\end{table}

Somehow surprisingly, when BN and ReLU are both used as pre-activation, the results are improved by healthy margins (Table~\ref{tab:activations} and Table~\ref{tab:preact}). In Table~\ref{tab:preact} we report results using various architectures: (i) ResNet-110, (ii) ResNet-164, (iii) a 110-layer ResNet architecture in which each shortcut skips only 1 layer (\ie, a Residual Unit has only 1 layer), denoted as ``ResNet-110(1layer)'', and (iv) a 1001-layer bottleneck architecture that has 333 Residual Units (111 on each feature map size), denoted as ``ResNet-1001''. We also experiment on CIFAR-100. Table~\ref{tab:preact} shows that our ``pre-activation'' models are consistently better than the baseline counterparts. We analyze these results in the following.

\subsection{Analysis}

We find the impact of pre-activation is twofold. First, the optimization is further eased (comparing with the baseline ResNet) because $f$ is an identity mapping. Second, using BN as pre-activation improves regularization of the models.

\textbf{Ease of optimization}. This effect is particularly obvious when training the \emph{1001-layer} ResNet. Fig.~\ref{fig:teaser} shows the curves. Using the original design in \cite{He2016}, the training error is reduced very slowly at the beginning of training. For $f=$ ReLU, the signal is impacted if it is negative, and when there are many Residual Units, this effect becomes prominent and Eqn.(\ref{eq:additive0}) (so Eqn.(\ref{eq:grad})) is not a good approximation. On the other hand, when $f$ is an identity mapping, the signal can be propagated directly between any two units. Our 1001-layer network reduces the training loss very quickly (Fig.~\ref{fig:teaser}). It also achieves the lowest loss among all models we investigated, suggesting the success of optimization.

We also find that the impact of $f=$ ReLU is not severe when the ResNet has fewer layers (\eg, 164 in Fig.~\ref{fig:curves_act}(right)). The training curve seems to suffer a little bit at the beginning of training, but goes into a healthy status soon. By monitoring the responses we observe that this is because after some training, the weights are adjusted into a status such that $\ve{y}_l$ in Eqn.(\ref{eq:resunit1}) is more frequently above zero and $f$ does not truncate it ($\ve{x}_l$ is always non-negative due to the previous ReLU, so $\ve{y}_l$ is below zero only when the magnitude of $\mathcal{F}$ is very negative). The truncation, however, is more frequent when there are 1000 layers.

\textbf{Reducing overfitting}. Another impact of using the proposed pre-activation unit is on regularization, as shown in Fig.~\ref{fig:curves_act} (right). The pre-activation version reaches slightly higher training loss at convergence, but produces lower test error. This phenomenon is observed on ResNet-110, ResNet-110(1-layer), and ResNet-164 on both CIFAR-10 and 100. This is presumably caused by BN's regularization effect \cite{Ioffe2015}. In the original Residual Unit (Fig.~\ref{fig:activations}(a)), although the BN normalizes the signal, this is soon added to the shortcut and thus the merged signal is not normalized. This unnormalized signal is then used as the input of the next weight layer. On the contrary, in our pre-activation version, the inputs to all weight layers have been normalized.

\hypersetup{urlcolor=magenta}

\section{Results}

\textbf{Comparisons on CIFAR-10/100.}
Table~\ref{tab:cifar} compares the state-of-the-art methods on CIFAR-10/100, where we achieve competitive results. We note that we do not specially tailor the network width or filter sizes, nor use regularization techniques (such as dropout) which are very effective for these small datasets. We obtain these results via a simple but essential concept --- going deeper.
These results demonstrate the potential of \emph{pushing the limits of depth}.

\renewcommand\arraystretch{1.33}
\setlength{\tabcolsep}{4pt}
\begin{table}[t]
\caption{Comparisons with state-of-the-art methods on CIFAR-10 and CIFAR-100 using ``\emph{moderate data augmentation}'' (flip/translation), except for ELU \cite{Clevert2016} with no augmentation. Better results of \cite{Graham2014,Springenberg2014} have been reported using stronger data augmentation and ensembling.
For the ResNets we also report the number of parameters.
Our results are the median of 5 runs with mean$\pm$std in the brackets. All ResNets results are obtained with a mini-batch size of 128 except $^\dag$ with a mini-batch size of 64 (code available at \url{https://github.com/KaimingHe/resnet-1k-layers}).}\label{tab:cifar}
\centering
\resizebox{!}{0.20\linewidth}{
\fontsize{8pt}{1em}\selectfont
\begin{tabular}{l|l}
\hline
  \textbf{CIFAR-10}   & error (\%) \\
\hline
NIN \cite{Lin2014a} & 8.81 \\
DSN \cite{Lee2015} & 8.22 \\
FitNet \cite{Romero2015} & 8.39 \\
Highway \cite{Srivastava2015a} & 7.72 \\
All-CNN \cite{Springenberg2014} & 7.25 \\
ELU \cite{Clevert2016} & 6.55 \\
FitResNet, LSUV \cite{Mishkin2016} & 5.84 \\
\hline
ResNet-110 \cite{He2016} (1.7M) & 6.61 \\
ResNet-1202 \cite{He2016} (19.4M) & 7.93 \\
\hline
ResNet-164 [ours] (1.7M) & 5.46 \\
ResNet-1001 [ours] (10.2M) & 4.92 \tiny{(4.89$\pm$0.14)} \\
ResNet-1001 [ours] (10.2M)$^\dag$ & \textbf{4.62} \tiny{(4.69$\pm$0.20)} \\
\hline
\end{tabular}
}
\resizebox{!}{0.20\linewidth}{
\fontsize{8pt}{1em}\selectfont
\begin{tabular}{l|l}
\hline
  \textbf{CIFAR-100}   & error (\%) \\
\hline
NIN \cite{Lin2014a} & 35.68 \\ 
DSN \cite{Lee2015} & 34.57 \\ 
FitNet \cite{Romero2015} & 35.04 \\ 
Highway \cite{Srivastava2015a} & 32.39 \\ 
All-CNN \cite{Springenberg2014} & 33.71 \\ 
ELU \cite{Clevert2016} & 24.28 \\ 
FitNet, LSUV \cite{Mishkin2016} & 27.66 \\ 
\hline
ResNet-164 \cite{He2016} (1.7M) & 25.16 \\
ResNet-1001 \cite{He2016} (10.2M) & 27.82 \\
\hline
ResNet-164 [ours] (1.7M) & 24.33 \\
ResNet-1001 [ours] (10.2M) & \textbf{22.71} {\tiny (22.68$\pm$0.22)} \\
\hline
\multicolumn{2}{c}{} \\
\end{tabular}
}
\end{table}

{\small ~\\}
\noindent\textbf{Comparisons on ImageNet.}
Next we report experimental results on the 1000-class ImageNet dataset \cite{Russakovsky2015}. We have done preliminary experiments using the skip connections studied in Fig.~\ref{fig:shortcuts} \& \ref{fig:curves_shortcuts}  on ImageNet with ResNet-101 \cite{He2016}, and observed similar optimization difficulties. The training error of these non-identity shortcut networks is obviously higher than the original ResNet at the first learning rate (similar to Fig.~\ref{fig:curves_shortcuts}), and we decided to halt training due to limited resources. But we did finish a ``BN after addition'' version (Fig.~\ref{fig:activations}(b)) of ResNet-101 on ImageNet and observed higher training loss and validation error. This model's single-crop (224$\times$224) validation error is 24.6\%/7.5\%, \vs the original ResNet-101's 23.6\%/7.1\%. This is in line with the results on CIFAR in Fig.~\ref{fig:curves_act} (left).

Table~\ref{tab:imagenet} shows the results of ResNet-152 \cite{He2016} and ResNet-200\footnote{The ResNet-200 has 16 more 3-layer bottleneck Residual Units than ResNet-152, which are added on the feature map of 28$\times$28.}, all trained from scratch. We notice that the original ResNet paper \cite{He2016} trained the models using scale jittering with shorter side $s\in[256, 480]$, and so the test of a 224$\times$224 crop on $s=256$ (as did in \cite{He2016}) is negatively biased. Instead, we test a single 320$\times$320 crop from $s=320$, for all original and our ResNets.
Even though the ResNets are trained on smaller crops, they can be easily tested on larger crops because the ResNets are fully convolutional by design. This size is also close to 299$\times$299 used by Inception v3 \cite{Szegedy2016a}, allowing a fairer comparison.

The original ResNet-152 \cite{He2016} has top-1 error of 21.3\% on a 320$\times$320 crop, and our pre-activation counterpart has 21.1\%. The gain is not big on ResNet-152 because this model has not shown severe generalization difficulties. However, the original ResNet-200 has an error rate of 21.8\%, higher than the baseline ResNet-152.
But we find that the original ResNet-200 has \emph{lower} training error than ResNet-152, suggesting that it suffers from overfitting.

Our pre-activation ResNet-200 has an error rate of 20.7\%, which is \textbf{1.1\%} lower than the baseline ResNet-200 and also lower than the two versions of ResNet-152. When using the scale and aspect ratio augmentation of \cite{Szegedy2015,Szegedy2016a}, our ResNet-200 has a result better than Inception v3 \cite{Szegedy2016a} (Table~\ref{tab:imagenet}). Concurrent with our work, an Inception-ResNet-v2 model \cite{Szegedy2016} achieves a single-crop result of 19.9\%/4.9\%. We expect our observations and the proposed Residual Unit will help this type and generally other types of ResNets.

\renewcommand\arraystretch{1.2}
\setlength{\tabcolsep}{3pt}
\begin{table}[t]
\caption{Comparisons of single-crop error on the ILSVRC 2012 validation set. All ResNets are trained using the same hyper-parameters and implementations as \cite{He2016}). Our Residual Units are the full pre-activation version (Fig.~\ref{fig:activations}(e)). $^\dag$: code/model available at \url{https://github.com/facebook/fb.resnet.torch/tree/master/pretrained}, using scale and aspect ratio augmentation in \cite{Szegedy2015}.
}\label{tab:imagenet}
\centering
\fontsize{8pt}{1em}\selectfont
\begin{tabular}{l|c|c|c|c|c}
\hline
 \tiny method & \tiny augmentation & \tiny train crop & \tiny test crop & \tiny top-1 & \tiny top-5 \\
\hline
ResNet-152, original Residual Unit \cite{He2016} & scale & 224$\times$224 & 224$\times$224 & 23.0  & 6.7 \\
\hline
ResNet-152, original Residual Unit \cite{He2016} & scale & 224$\times$224 & 320$\times$320 & 21.3  & 5.5 \\
ResNet-152, \textbf{pre-act} Residual Unit & scale & 224$\times$224 & 320$\times$320 & 21.1 & 5.5 \\
\hline
ResNet-200, original Residual Unit \cite{He2016} & scale & 224$\times$224 & 320$\times$320 & 21.8 & 6.0 \\
ResNet-200, \textbf{pre-act} Residual Unit & scale & 224$\times$224 & 320$\times$320 & \textbf{20.7} & \textbf{5.3} \\
ResNet-200, \textbf{pre-act} Residual Unit & scale+asp ratio & 224$\times$224 & 320$\times$320 & \textbf{20.1}$^\dag$ & \textbf{4.8}$^\dag$ \\
\hline
Inception v3 \cite{Szegedy2016a} & scale+asp ratio & 299$\times$299 & 299$\times$299 & 21.2 & 5.6 \\
\hline
\end{tabular}
\end{table}

{\small ~\\}
\noindent\textbf{Computational Cost.}
Our models' computational complexity is linear on depth (so a 1001-layer net is $\sim$10$\times$ complex of a 100-layer net). On CIFAR, ResNet-1001 takes about 27 hours to train on 2 GPUs; on ImageNet, ResNet-200 takes about 3 weeks to train on 8 GPUs (on par with VGG nets \cite{Simonyan2015}).

\section{Conclusions}

This paper investigates the propagation formulations behind the connection mechanisms of deep residual networks. Our derivations imply that identity shortcut connections and identity after-addition activation are essential for making information propagation smooth. Ablation experiments demonstrate phenomena that are consistent with our derivations. We also present 1000-layer deep networks that can be easily trained and achieve improved accuracy.

\subsubsection*{Appendix: Implementation Details}

The implementation details and hyper-parameters are the same as those in \cite{He2016}. On CIFAR we use only the translation and flipping augmentation in \cite{He2016} for training. The learning rate starts from 0.1, and is divided by 10 at 32k and 48k iterations. Following \cite{He2016}, for all CIFAR experiments we warm up the training by using a smaller learning rate of 0.01 at the beginning 400 iterations and go back to 0.1 after that, although we remark that this is not necessary for our proposed Residual Unit. The mini-batch size is 128 on 2 GPUs (64 each), the weight decay is 0.0001, the momentum is 0.9, and the weights are initialized as in \cite{He2015}.

On ImageNet, we train the models using the same data augmentation as in \cite{He2016}. The learning rate starts from 0.1 (no warming up), and is divided by 10 at 30 and 60 epochs. The mini-batch size is 256 on 8 GPUs (32 each). The weight decay, momentum, and weight initialization are the same as above.

When using the pre-activation Residual Units (Fig.~\ref{fig:activations}(d)(e) and Fig.~\ref{fig:asymmetric}), we pay special attention to the first and the last Residual Units of the entire network. For the first Residual Unit (that follows a stand-alone convolutional layer, conv$_1$), we adopt the first activation right after conv$_1$ and before splitting into two paths; for the last Residual Unit (followed by average pooling and a fully-connected classifier), we adopt an extra activation right after its element-wise addition.
These two special cases are the natural outcome when we obtain the pre-activation network via the modification procedure as shown in Fig.~\ref{fig:asymmetric}.

\newcommand{\blocka}[1]{\tiny \begin{tabular}{c} \(\left[\begin{array}{c}\text{3$\times$3, #1}\\ \text{3$\times$3, #1} \end{array}\right]\)\end{tabular}
}
\newcommand{\blockb}[2]{\tiny \begin{tabular}{c} \(\left[\begin{array}{c}\text{1$\times$1, #2}\\[-.1em] \text{3$\times$3, #2}\\[-.1em] \text{1$\times$1, #1}\end{array}\right]\)\end{tabular}
}

The bottleneck Residual Units (for ResNet-164/1001 on CIFAR) are constructed following \cite{He2016}. For example, a $\blocka{16}$ unit in ResNet-110 is replaced with a $\blockb{64}{16}$ unit in ResNet-164, both of which have roughly the same number of parameters. For the bottleneck ResNets, when reducing the feature map size we use projection shortcuts \cite{He2016} for increasing dimensions, and when pre-activation is used, these projection shortcuts are also with pre-activation.

\bibliographystyle{splncs}
\bibliography{resnet_plus_arxiv}

\end{document}